\documentclass[letterpaper, 10pt, conference]{ieeeconf} 

\IEEEoverridecommandlockouts       
\usepackage{times} 
\usepackage{graphicx} 
\usepackage{bm} 
\usepackage{amsmath} 
\usepackage{amssymb} 
\usepackage{color} 
\usepackage[ruled,linesnumbered]{algorithm2e} 
\usepackage[short]{optidef} 
\usepackage{gensymb} 
\usepackage{multirow}
\usepackage[mathscr]{eucal}
\usepackage{threeparttable}

\usepackage{xparse,soul} 
\soulregister\cite7 
\soulregister\citep7 
\soulregister\citet7 
\soulregister\ref7 
\soulregister\pageref7 
\usepackage{makecell}

\usepackage{fancyhdr}
\fancypagestyle{firstpage}{
    \chead{This paper has been accepted for publication at the 2025 International Conference on Intelligent Robots and Systems.}}
\usepackage{cite}
\title{S$^{3}$D: A Spatial Steerable Surgical  Drilling  Framework for  Robotic Spinal Fixation Procedures}
\author{Daniyal Maroufi$^{*1}$, Xinyuan Huang$^{*1}$, Yash Kulkarni$^{*1}$, Omid Rezayof$^{1}$, Susheela Sharma$^{1}$, Vaibhav Goggela$^{1}$, \\ Jordan P. Amadio$^{2}$, Mohsen Khadem$^{3}$, and Farshid Alambeigi$^{1}$ \IEEEmembership{Member, IEEE}
\thanks{*These authors contributed equally to this work.}
\thanks{**This work was supported in part by the National Institute Of Biomedical Imaging and Bioengineering of the National Institutes of Health under Award Number R21EB030796 and in part by the Collaborative Accelerator for Transformative Research Endeavors grant, jointly awarded by The University of Texas at Austin and The University of Texas MD Anderson Cancer Center.}
\thanks{$^{1}$D.~Maroufi, X.~Huang, Y.~Kulkarni, O.~Rezayof, S.~Sharma, V. ~Goggela and F.~Alambeigi are with the Walker Department of Mechanical Engineering and the Texas Robotics  at the University of Texas at Austin, Austin, TX, 78712, USA. Email: \{maroufi, hxydiana, kulkarni.yash08, omid.rezayof, sheela.sharma\}@utexas.edu,   \{farshid.alambeigi\}@austin.utexas.edu}.
\thanks{$^{2}$J.~P.~ Amadio is with the Department of Neurosurgery, The University of Texas Dell Medical School, TX, 78712. }
\thanks{$^{3}$M.~Khadem is with the School of Informatics, University of Edinburgh. }}

\usepackage{color} 

\begin{document}
\maketitle
\thispagestyle{firstpage}
\pagestyle{empty}
		
\begin{abstract}
In this paper, we introduce S$^{3}$D: A Spatial Steerable Surgical  Drilling  Framework for  Robotic Spinal Fixation Procedures. S$^{3}$D is designed to enable realistic steerable drilling while accounting for the anatomical constraints associated with vertebral access in spinal fixation (SF) procedures. To achieve this, we first enhanced our previously designed  concentric tube Steerable Drilling Robot (CT-SDR) to facilitate steerable drilling across all vertebral levels of the spinal column.  Additionally, we propose a four-Phase calibration, registration, and navigation procedure to perform realistic SF procedures on a spine holder phantom by integrating  the CT-SDR with a seven-degree-of-freedom robotic manipulator. The functionality of this framework is validated through planar and out-of-plane steerable drilling experiments in vertebral phantoms.
\end{abstract}

\section{Introduction}

Spinal Fixation (SF) is one of the most common surgical procedures performed by spine and orthopedic surgeons as a treatment for a broad variety of spinal disorders including vertebral compression fractures, spinal tumors that require vertebral reconstruction, and degenerative spinal diseases \cite{gaines2000use}.
As the gold standard approach for SF, Pedicle Screw Fixation (PSF) is a minimally invasive, image-guided surgical procedure that utilizes a rigid drilling instrument to create a linear tunnel through the pedicle of the vertebrae. A Rigid Pedicle Screw (RPS) is then inserted into this tunnel to reach the  cancellous bone region of the vertebral body, and fixate vertebrae \cite{rometsch2020screw}.
Despite the benefits of PSF, the rigidity of surgical instruments and pedicle screws, combined with the complex anatomy of the vertebrae, presents several challenges for such procedures. First, achieving optimal angles for drilling and screw placement is impeded by the lack of instrument steerability and limited access to the vertebral body. This issue is particularly significant in cases of osteoporotic vertebrae, where insufficient Bone Mineral Density (BMD) increases the risk of fixation failure and loosening or pullout of the pedicle screw \cite{wittenberg1991importance,okuyama1993stability,weiser2017insufficient}.
In addition, the complexity of this procedure requires highly skilled and experienced surgeons. Furthermore, restricted maneuverability of instruments around vital anatomical structures can lead to neurovascular compromise and injury to the nerve root and/or spine, as well as screw misplacement and failure of fixation.

\begin{figure}[t]
   \centering
   \includegraphics[width=1.0\linewidth]{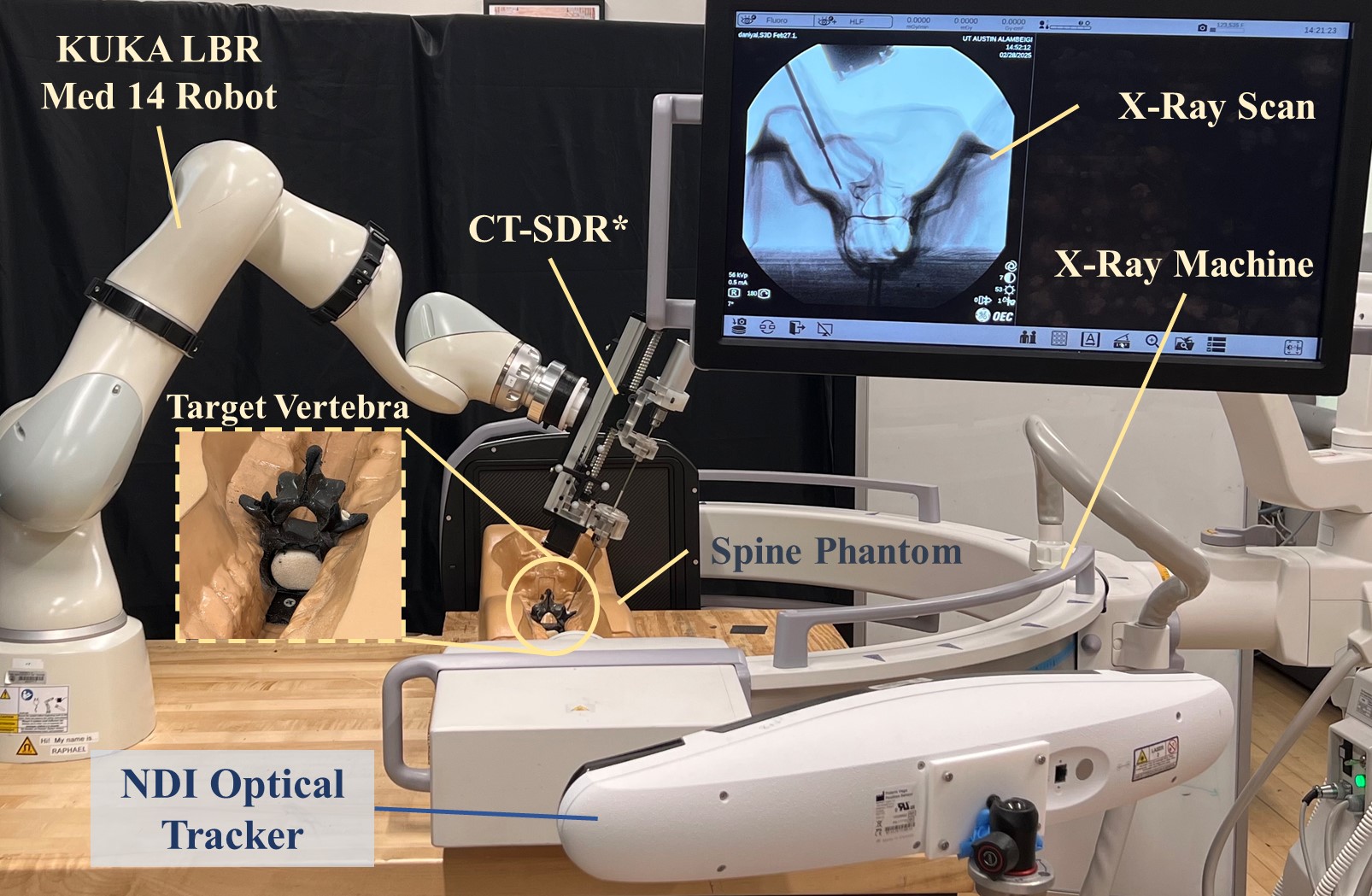}
   \caption{ The experimental setup consisting of a 7 DoF KUKA robotic arm, the CT-SDR$^{*}$, a C-arm X-ray machine, the spine phantom, and NDI optical tracking system. The target vertebra is shown in a zoomed-in view. An X-ray scan of the pilot hole drilling stage is also shown on the X-ray machine's display monitor. 
   }
   \label{fig:setup}
\end{figure}

To partly overcome these challenges, several navigation and robotic technologies have been developed \cite{Li2023RoboticSA}. 
Surgical navigation systems make use of preoperative imaging techniques, such as CT scans, and/or intraoperative imaging methods, like fluoroscopy, to construct a comprehensive 3D model of the patient's vertebrae \cite{CT_MRI_registration,stent_recovery}.
By employing optical tracking systems, surgeons can track instruments with respect to this 3D model in real time. This functionality improves the accuracy of drilling and screw placement, reducing the risk of screw misplacement and possible nerve damage \cite{Li2023RoboticSA,ElmiTerander2018PedicleSP}. However, the success of the process remains highly dependent on the clinician's skill in conducting it. To address this issue, commercially available surgical robotic systems such as ExcelsiusGPS (Globus Medical, PA, USA) and Mazor X (Medtronic) assist the surgeon in holding and guiding the tools for more precise drilling and pedicle screw placement \cite{beyond2025Bhimreddy}.
Although these systems offer advantages, the use of rigid tools and pedicle screws leads to the same issues as manual methods, including limited access to regions with high BMD. Furthermore, inflexibility of instruments and the anatomical constraints of the vertebrae limit the available workspace, making effective surgical planning unfeasible in these systems.

Recently several Steerable Drilling Robots (SDRs) have been introduced to target difficult-to-access areas within complex anatomical structures, including spine and pelvis, facilitating implant placement in high BMD regions \cite{wang2021design,9732206,alambeigi2019use,alambeigi2017curved}. In particular, our group has introduced Concentric Tube Steerable Drilling Robots (CT-SDRs) as an intuitive and robust solution for executing various C-, J-, U- and S-shaped trajectories in spinal fixation (SF) procedures \cite{sharma2024patient,Sharma_tbme_2022,Sharma_ismr, Kulkarni2025TowardsDD} along with Flexible Pedicle Screws (FPS) capable of following these curved trajectories \cite{Kulkarni2025SynergisticPSA,Kulkarni2024SFF, Kulkarni2025AugmentedBSF}. While our CT-SDR developments mark significant progress, several critical limitations hinder the real-world applicability of the proposed system. Specifically, \textbf{(i)} The CT-SDR design presented in \cite{Sharma_tbme_2022,sharma2023novel} is constrained to guiding tubes and instruments of specific diameters, thereby restricting its usability to a limited range of vertebral levels. This lack of adaptability reduces the system’s generalizability for SF procedures; \textbf{(ii)} The calibration, registration, and semi/autonomous robotic-assisted procedures proposed in \cite{sharma2024patient,sharma2024biomechanics} have been validated using simulated bone blocks or single-level vertebrae. However, these studies do not account for the actual patient anatomy or the precise approach angle required for pedicle entry. As a result, the experiments were conducted with fixed and unrealistic pedicle approach angles; and \textbf{(iii)}
The drilling experiments were performed exclusively on Sawbone phantoms, which solely simulate the cancellous bone of the vertebral body. In a real SF procedure, however, the drill must first penetrate the cortical bone of the pedicle before advancing into the cancellous bone of the vertebral body, a critical aspect not addressed in these studies further limiting applicability of these studies for a real surgical scenario.

To collectively address the above-mentioned limitations and as our main contributions, in this paper, we introduce S$^{3}$D: A Spatial Steerable Surgical  Drilling  Framework for  Robotic Spinal Fixation Procedures. S$^{3}$D is designed to enable realistic steerable drilling while accounting for the anatomical constraints associated with vertebral access in SF procedures. To achieve this, we first enhanced our previously designed  CT-SDR \cite{Sharma_tbme_2022} to facilitate steerable drilling across all vertebral levels of the spinal column. Of note, in this paper, we call this modified design as CT-SDR$^{*}$. Additionally, we propose a four-Phase calibration, registration, and navigation procedure to perform realistic SF procedures on a spine holder phantom by integrating  the CT-SDR$^{*}$ with a seven-degree-of-freedom (DOF) robotic manipulator. The functionality of this framework is validated through steerable drilling experiments on vertebral phantoms, simulating both cortical and cancelous bone in an osteoporotic patient model.

\section{Autonomous Drilling Robotic System} \label{sec:method}

As shown in Fig. \ref{fig:setup}, the S$^{3}$D framework is composed of  a 7 DoF robotic arm (KUKA lbr-Med 14, KUKA, Germany), an optical tracking system (NDI Polaris Vega, Northern Digital Inc.),  CT-SDR$^{*}$, and complementary flexible drilling instruments.
The robotic arm is used to accurately move and place the CT-SDR$^{*}$ to desired drilling poses (i.e., spatial position and orientation), and to perform the insertion and retraction of the drill for performing a SF procedure. The optical tracking system is employed to measure the desired drilling poses in real-time and to aid hardware calibration and registration, which will be explained in the later sections.  
Lastly, the CT-SDR$^{*}$ is designed to first drill a straight pilot hole inside the cortical bone of vertebral pedicle and subsequently drill J-shaped trajectories inside the cancelous bone of vertebral body. The details of the newly designed CT-SDR$^{*}$ are outlined below. 

\subsection{Concentric Tube Steerable Drilling Robot$^{*}$ (CT-SDR$^{*}$)} \label{subsec:CTSDR}
\begin{figure}[t]
   \centering
   \includegraphics[width=1.0\linewidth]{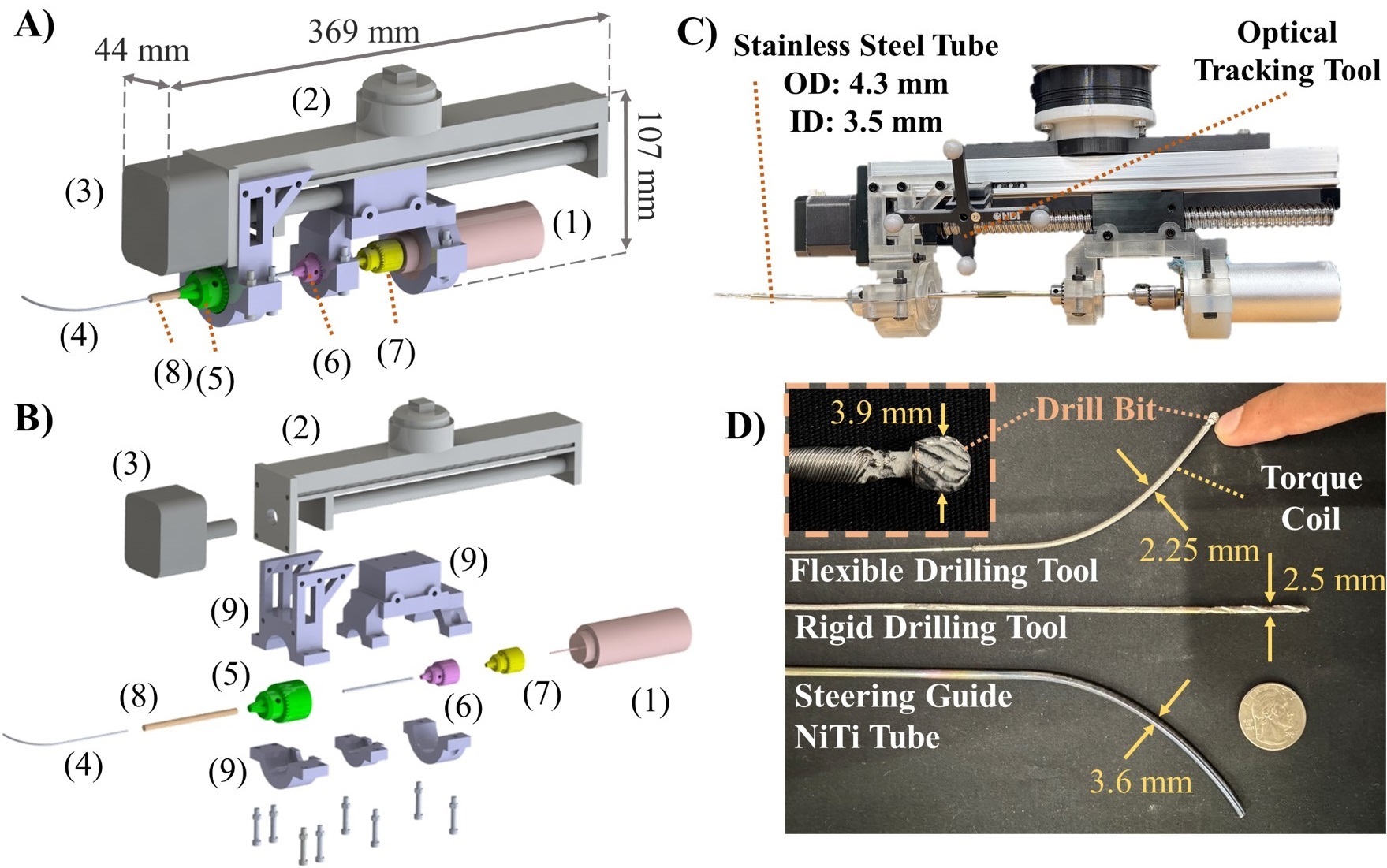}
   \vspace*{-6mm}
   \caption{ CT-SDR$^{*}$ Design. A,B) Rendered figures of assembled and exploded view of the CT-SDR$^{*}$, The marked components are:(1) Drilling Motor, (2) linear stage, (3) stepper motor, (4) steering guide tube, (5-7) adjustable chucks, (8) stainless steel tube, (9) CT-SDR$^{*}$ structural components C) The fabricated CT-SDR$^{*}$. An optical tracking tool is attached to the CT-SDR$^{*}$ body enabling the motion tracking by the optical tracker. D) Drilling tools, including the rigid and flexible drilling tools, steering guide and a zoomed-in view of the flexible drilling tool drill bit. 
   }
   \label{fig:drill}
\end{figure}

Our previously introduced CT-SDR system was extensively evaluated for its ability to generate safe and reliable curved trajectories spanning the entire vertebral body \cite{Sharma_ismr,Sharma_tbme_2022}. These capabilities were achieved through three key components: (i) a flexible cutting tool that enabled precise drilling into bone tissue, (ii) a reliable steering guide that ensured accurate drilling along a surgeon-specified path, and (iii) actuation units that provided controlled insertion, retraction, and rotational torque for drilling.
Building on this foundation, we present an enhanced CT-SDR$^{*}$ system designed to accommodate  steerable drilling across all vertebral levels of the spinal column by (i) using a wider range of steering guides and (ii)  working with the existing rigid surgical drill bits.  These advancements are crucial for expanding surgical access to previously inaccessible vertebral bodies (i.e., cervical spine) using CT-SDR. The rigid tool is also vital for enabling drilling into harder cortical bone.
\begin{figure*}[t!] 
	\centering 
	\includegraphics[width=0.85\linewidth]{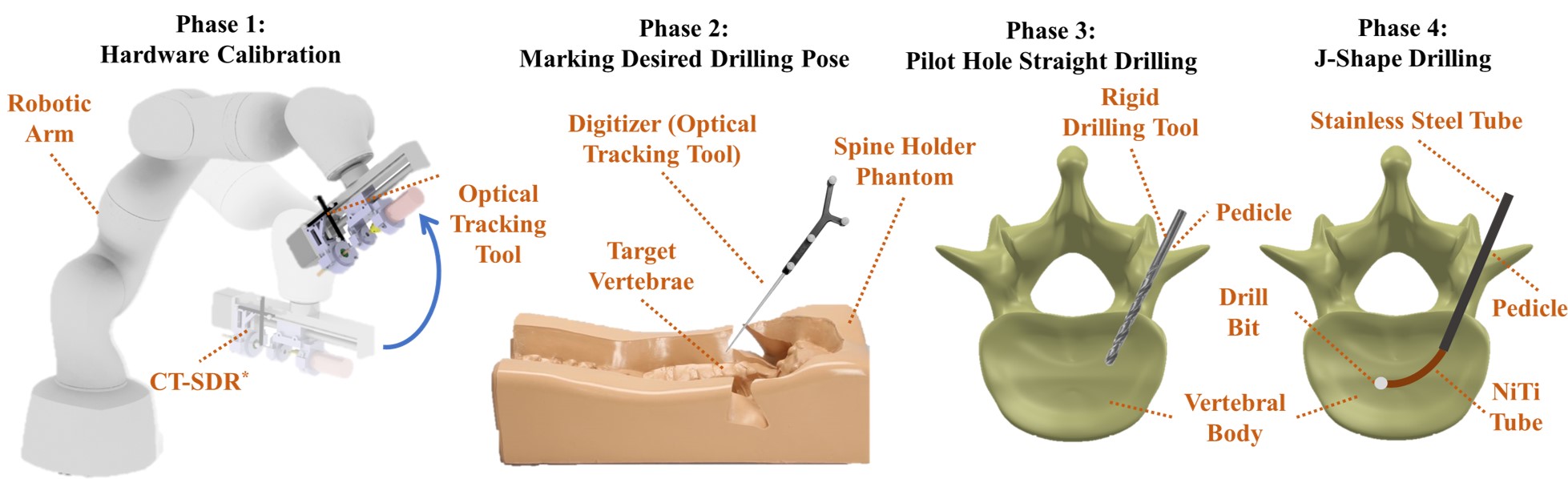}
        \vspace*{-3mm}
	\caption{The proposed four-Phase calibration, registration, and navigation procedure for performing an autonomous  realistic J-shape drilling. }
	\label{fig:framework}  
\end{figure*}
To meet these new requirements, we redesigned the CT-SDR with an adjustable chuck mechanism, allowing for compatibility with different steering guides and facilitating tool interchangeability by quick replacement of drilling instruments with different diameters  (Fig. \ref{fig:drill}). This updated design enables seamless transitions between drilling tool for cancelous and cortical bone to robustly create J-shaped trajectories. 

\subsubsection{Flexible Drilling Tool}
The flexible cutting tools used for the CT-SDR$^{*}$ are shown in Fig. \ref{fig:drill}D \cite{Sharma_ismr,Sharma_tbme_2022}. These tools were created by laser welding a 3.9 mm ball head burr (8175A29, McMaster-Carr) to a 104 mm long torque coil (Asahi Intecc. USA, Inc.). 
Finally, the torque coil was laser welded to a stainless steel tube (Mcmaster-Carr) for extending its length.
A ball head burr was chosen as the drill tip due to its ability to smoothly create curved trajectories. The 3.9 mm diameter drill bit is shown in \ref{fig:drill}D. A 2.25 mm outer diameter (OD) torque coil is the main component to ensure rotational motion is reliably delivered from the drill chuck (B075SZZN4J, Amazon) to the ball head burr, with the 1.3 mm stainless steel tube providing a stable interface between the flexible tool and the drill chuck. 

\subsubsection{Rigid Drilling Tool} To enable a robust drilling procedure inside cortical bone of vertebral pedicle, we integrated a commercial  rigid surgical drill (Sas interational, India) with CT-SDR$^{*}$. This new drilling tool works seamless with our novel design while providing us with a structurally stiffer drilling instrument to potentially drill into cortical bone. Towards this, as shown in Fig. \ref{fig:drill}D, a rigid drilling tool was fabricated by laser welding  a 2.5 mm surgical drill bit with a stainless steel tube (Mcmaster-Carr).  The dimensions of the rigid drilling tool can be seen in Fig. \ref{fig:drill}D.
\subsubsection{Steering Guide} Building on the work in our previous studies \cite{Sharma_ismr,Sharma_tbme_2022}, biocompatible Nitinol (NiTi) tubes were used to guide the flexible drilling tool during the insertion process. Due to the NiTi's super elasticity, shape memory properties, and structural integrity, it remains the optimal material for achieving the surgeon-specified curved trajectories. In this study, without loss of generality, we used a pre-curved NiTi tube with a 69.5 radius of curvature with 3.6 mm and 3 mm outer and inner diameters. The steerable guide NiTi tube is also depicted in Fig. \ref{fig:drill}D. 
\subsubsection{Adjustable Drill Chucks} 
During surgical procedures, the NiTi tube plays a crucial role in enabling the CT-SDR$^*$ system to achieve various radii of curvature. Smaller NiTi tubes, which can bend to a greater radius, enhance the system's ability to access smaller vertebral bodies, such as those in the cervical spine. To support this functionality, we have developed our CT-SDR$^*$ system with three adjustable chucks positioned at the beginning, middle, and end of the drill (chucks 5, 6, and 7, as shown in Fig. \ref{fig:drill}A). These adjustable chucks allow for the use of NiTi steering guides of various diameters, enabling the surgeon to select and readily use the most appropriate guide based on the patient's specific anatomical requirements. Specifically, chuck 5 is capable of housing drill shanks ranging from 1 to 10 mm (2812A38, McMaster-Carr), while chuck 6 can accommodate drill shanks from 0.4 to 4 mm (2812A17, McMaster-Carr). The drill chuck is also capable of accommodating drill shanks ranging from 0.3 to 4 mm. 
\subsubsection{Linear Actuation Units} 
The insertion movement of the drilling tools was controlled by an actuation unit consisting of a linear stage and a NEMA 17 stepper motor controlled by an Arduino Nano \cite{Sharma_ismr,Sharma_tbme_2022}.
The rotational speed of the drilling tool was controlled using a manual controller provided with the drill. This combination is vital for ensuring the drilling tools perform safe and reliable drilling. The linear stage and the  stepper motor can be seen in Fig. \ref{fig:drill}A and C.

\subsection{Autonomous Drilling Framework}

The proposed four-Phase calibration, registration, and
navigation procedure for performing an autonomous   realistic J-shape drilling is depicted in Fig. \ref{fig:framework}. As shown, the framework consists of four main phases, including: (Phase 1) hardware calibration, (Phase 2) pose marking, (Phase 3) pilot hole drilling, and (Phase 4) J-shape drilling. 
In Phase 1,  the unknown transformations between different frames including the tip of the CT-SDR$^*$ with respect to the robot end-effector are found. 
In Phase 2, the surgeon defines the desired entry point and orientation to the vertebral pedicle on the vertebra using an optical digitizer. 
In Phase 3, the obtained pose in Phase 2 is  used to move the CT-SDR$^*$ drilling instrument to the marked entry point with the same orientation. 
  Next, the CT-SDR$^*$ creates a straight pilot hole inside the pedicle  using the rigid surgical drilling tool. 
Finally and in Phase 4, the rigid drilling tool is substituted with flexible drilling tool and CT-SDR$^*$ is used to create a J-shape   drilling trajectory. 
Once the drilling is complete, the robot is instructed to return to the home position. The following sections will provide a detailed explanation of each step of the proposed framework.

\subsubsection{\textbf{Phase 1:} Hardware Calibration} \label{subsubsec:Phase1-calibration}
\begin{figure}[t!] 
	\centering 
	\includegraphics[width=0.95\linewidth]{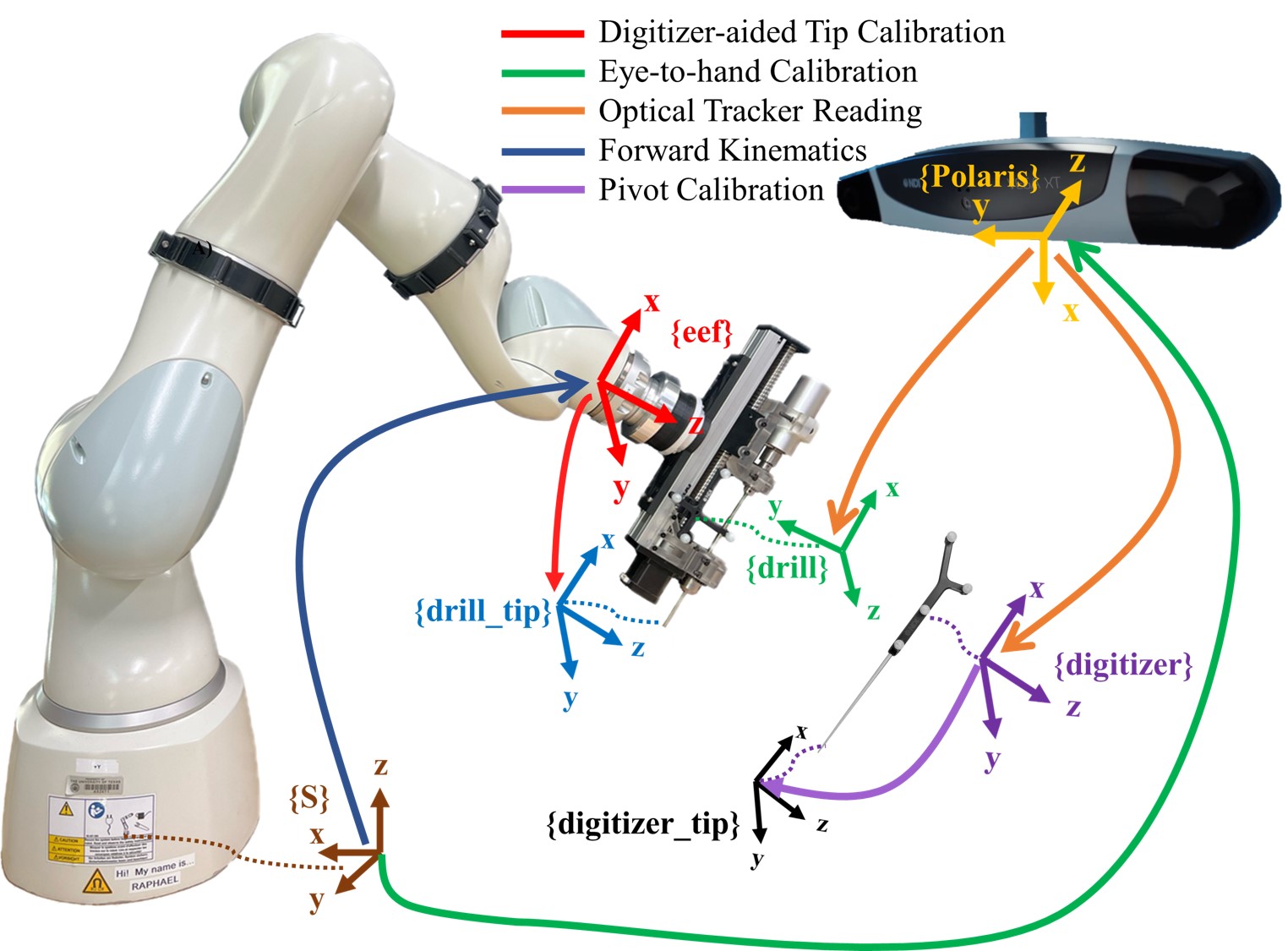}
	\caption{Frames assigned to points-of-interest in the system along with the employed calibration methods to identify missing transformations between the frames.}
	\label{fig:calibrations}
\end{figure}

The calibration and registration phase uses three algorithms to find the unknown transformations required in autonomous placement of the CT-SDR$^*$, including: hand-eye calibration, pivot calibration and digitizer-aided tip calibration. 
As shown in Fig. \ref{fig:calibrations}, the autonomous drilling framework utilizes multiple coordinate frames and counts on relevant transformations between these coordinate frames to precisely move the robot arm.
While some of these transformations can be directly measured (e.g. by the the optical tracker), others are missing and require addition calibration efforts to determine.
The details of these calibration methods are provided in the following sections. 

{\textbf{Hand-Eye Calibration:} }
A hand-eye calibration algorithm is employed to determine the transformation from the optical tracker to the robot's base frame (i.e., world frame or $\{S\}$). 
The transformation from robot's end-effector to the optical tracking tool attached on the CT-SDR$^{*}$, also known as $\{\text{drill}\}$,   is also found as a result of this calibration.
We formulate this as an $AX=ZB$ problem and solve with Kronecker product solution \cite{shah2013solving}, implemented in OpenCV library \cite{opencv_library}.
The system of equations can be written as:

\begin{equation}\label{eq:axzb}
\begin{bmatrix} ... \\ {}^{\text{drill}}T_{\text{Polaris}}^{i} \\ ... \end{bmatrix} 
\begin{bmatrix} X \end{bmatrix} =
\begin{bmatrix} Z \end{bmatrix} 
\begin{bmatrix} ... \\ {}^{\text{eef}}T_{S}^{i} \\ ... \end{bmatrix}
\end{equation}
where ${}^{\text{drill}}T_{\text{Polaris}} \in SE(3)$ represents the transformation of the optical tracker frame with respect to the drill frame, ${}^{\text{eef}}T_{S} \in SE(3)$ denotes the transformation of the world frame in relation to the robot's end effector frame, and $i$ represents the $i$th pose captured.

The solution yields the two missing spatial transformations: $X \in SE(3)$, also known as ${}^{\text{Polaris}}T_{S}$, whose inverse describes the transformation from the optical tracker to the world frame, and $Z \in SE(3)$, also known as ${}^{\text{drill}}T_{\text{eef}}$, which represents the transformation from the robot's end effector with respect to the optical tracking tool. With the result of hand-eye calibration, two major coordinate frame systems, the robot and the optical tracker, can now be associated.

{\textbf{Pivot Calibration:}}
To precisely determine the position offset between the optical digitizer's tool tip and its frame, 
a pivot calibration algorithm \cite{yaniv2015pivot} was employed. 
The missing translation is calculated by rotating the digitizer around a fixed pivot point while recording its transform with respect to the optical tracking system $\{\text{Polaris}\}$. For each configuration $i$, the pivot point's location \(x_{\text{pivot}} \in \mathbb{R}^3\) with respect to $\{{\text{Polaris}\}}$, can be computed using
$
x_{\text{pivot}} = R_i x_{\text{tip}} + p_i
$
where \(p_i \in \mathbb{R}^3\) and \(R_i \in SO(3)\)  represent the digitizer's position and orientation in $\{\text{Polaris}\}$, respectively. The offset distance to the digitizer's tip, \(x_{\text{tip}} \in \mathbb{R}^3\) with respect to digitizer's frame, is determined by solving an overdetermined system of equations:
\begin{equation}\label{eq:pivot_eq}
\begin{bmatrix} 
\vdots & \vdots \\ 
R_i & -I \\ 
\vdots & \vdots 
\end{bmatrix} 
\begin{bmatrix} 
x_{\text{tip}} \\ 
x_{\text{pivot}} 
\end{bmatrix}
= 
\begin{bmatrix} 
\vdots \\ 
-p_i \\ 
\vdots 
\end{bmatrix}
\end{equation}
using least-squares fitting, where \(i\) refers to the \(i\)th pose captured. 

{\textbf{Digitizer-aided Tip Calibration:}}
In this calibration,  digitizer tool is used to point at the drill tip (without applying excessive force to move the drill tip) while holding different orientations. The digitizer's pose with respect to the optical tracking system is then acquired. Using  transformations acquired from previous calibration sections, the digitizer tip's pose with respect to the optical tracking system can be calculated as:
\begin{equation}\label{eq:digi_wrt_s_eq}
{}^{S}T_{\text{digitizer}} =
\begin{bmatrix} 
{}^{S}R_{\text{digitizer}} & {}^{S}x_{\text{digitizer}}\\ 
\boldsymbol{0} & 1
\end{bmatrix} 
 = {}^{S}T_{\text{Polaris}} {}^{\text{Polaris}}T_{\text{digitizer}}
\end{equation}

\begin{equation}\label{eq:digitip_wrt_s_eq}
t_i = {}^{S}x_{\text{digitizer\_tip}} = {}^{S}R_{\text{digitizer}} {}^{\text{digitizer}}x_{\text{digitizer\_tip}} + {}^{S}x_{\text{digitizer}}
\end{equation}
where transformations \({}^{S}T_{\text{digitizer}}, {}^{S}T_{\text{Polaris}}, {}^{\text{Polaris}}T_{\text{digitizer}} \in SE(3) \), rotation matrices \( {}^{S}R_{\text{digitizer}} \in SO(3) \), and vectors \( {}^{S}x_{\text{digitizer}}, {}^{S}x_{\text{digitizer\_tip}}, {}^{\text{digitizer}}x_{\text{digitizer\_tip}}, {}^{S}x_{\text{digitizer}} \in \mathbb{R}^3 \).
Lastly, we solve the following linear system of equations using least-squares fitting:
\begin{equation}\label{eq:digitizer_aided_tip_calib_eq}
\begin{bmatrix} 
\vdots\\ 
R_i \\ 
\vdots 
\end{bmatrix} 
x_{\text{drill\_tip}}
= 
\begin{bmatrix} 
\vdots \\ 
t_i-p_i \\ 
\vdots 
\end{bmatrix}
\end{equation}
where \(R_i \in SO(3)\) and \(p_i \in \mathbb{R}^3\) represent the robot's end effector orientation and position in $\{S\}$, respectively, \(t_i \in \mathbb{R}^3\) represents the position of the digitizer tip in $\{S\}$, and $i$ refers to the $i$th pose captured.

\subsubsection{\textbf{Phase 2:} Marking Desired Drilling Pose} \label{subsubsec:phase2}

Upon completion of system calibration,  a digitizer, 
depicted in Fig. \ref{fig:framework}, is used to mark the desired entry point and drilling orientation for the CT-SDR$^*$. Specifically, a surgeon can point to a specific location on the pedicle of the vertebra with the digitizer tip, and align the digitizer shaft with the desired drilling direction for the initial straight part of the drilling trajectory.
The digitizer's position and orientation, represented by attached markers, is read by the optical tracker as ${}^{\text{Polaris}}T_{\text{digitizer}}$. 

To command the drill tip of the CT-SDR$^*$ to the desired position identified by the digitizer tip and orient the CT-SDR$^*$ to the desired drilling orientation, we improve upon the method used in \cite{sharma2024patient}. Previously, the digitizer was used to mark a desired position for the drilling, however, the orientation was not specified by the digitizer. Here, the digitizer orientation is further used to adjust the CT-SDR$^*$ orientation when drilling.  
Specifically, the desired pose of drill tip with respect to the optical tracking system is obtained by Eq. \ref{eq:T_eef_wrt_Polaris} which is based on the digitizer orientation when marking the desired drilling pose.
\begin{equation}
\label{eq:T_eef_wrt_Polaris}
{}^{\text{Polaris}}T_{\text{drill\_tip}}^{\small\textit{desired}} = {}^{\text{Polaris}}T_{\text{digitizer}} {}^{\text{digitizer}}T_{\text{digitizer\_tip}}
\end{equation}

where ${}^{\text{digitizer}}T_{\text{digitizer\_tip}}$ can be acquired by pivot calibration or NDI's product specification. Then the desired pose of robot end-effector can be calculated by following a series of frame transformations:
\begin{equation}\label{eq:T_eef_wrt_S}
{}^{S}T_{\text{eef}}^{\small\textit{desired}} = {}^{S}T_{\text{Polaris}} {}^{\text{Polaris}}T_{\text{drill\_tip}}^{\small\textit{desired}} {}^{\text{drill\_tip}}T_{\text{eef}} 
\end{equation}
where ${}^{S}T_{\text{Polaris}}$ and the translation offset in ${}^{\text{drill\_tip}}T_{\text{eef}}$ are identified in Section \ref{subsubsec:Phase1-calibration}.

\subsubsection{\textbf{Phase 3:} Pilot Hole Straight Drilling}
After the desired drilling pose is marked and recorded using the digitizer, the rigid drilling tool is inserted into the CT-SDR$^*$. 
The robot then moves the drill to the desired drilling pose and drilling is performed autonomously with robot controlling the insertion. As a result, a straight pilot hole is created through the pedicle and the cortical bone of the vertebra. Upon completion, robot retracts the CT-SDR$^*$ out of the pilot hole.

\subsubsection{\textbf{Phase 4:} J-Shape Drilling} 
\label{subsubsec:phase4}
To prepare for J-shape drilling, the robot moves the CT-SDR$^*$ to maintenance position, allowing the operator to replace the rigid drilling tool with the flexible drilling tool. After that, the robot moves the CT-SDR$^*$ back to the desired drilling pose so that the flexible drill tip is at previously marked drill entry point. The flexible drilling tool will first drill through the pedicle along the pilot hole created in the previous phase. In this step robot arm controls the insertion of the drill. Subsequently, J-shape drilling starts and is controlled by the linear actuation mechanism in the CT-SDR$^*$. The curved trajectory is guided by the NiTi tube which gradually moves out of the containing stainless tube. After drilling is done, the drill is retracted and then moved back to home position.

\section{Experimental Evaluations}
\begin{figure*}[t!] 
	\centering 
    \includegraphics[width=0.85\linewidth]{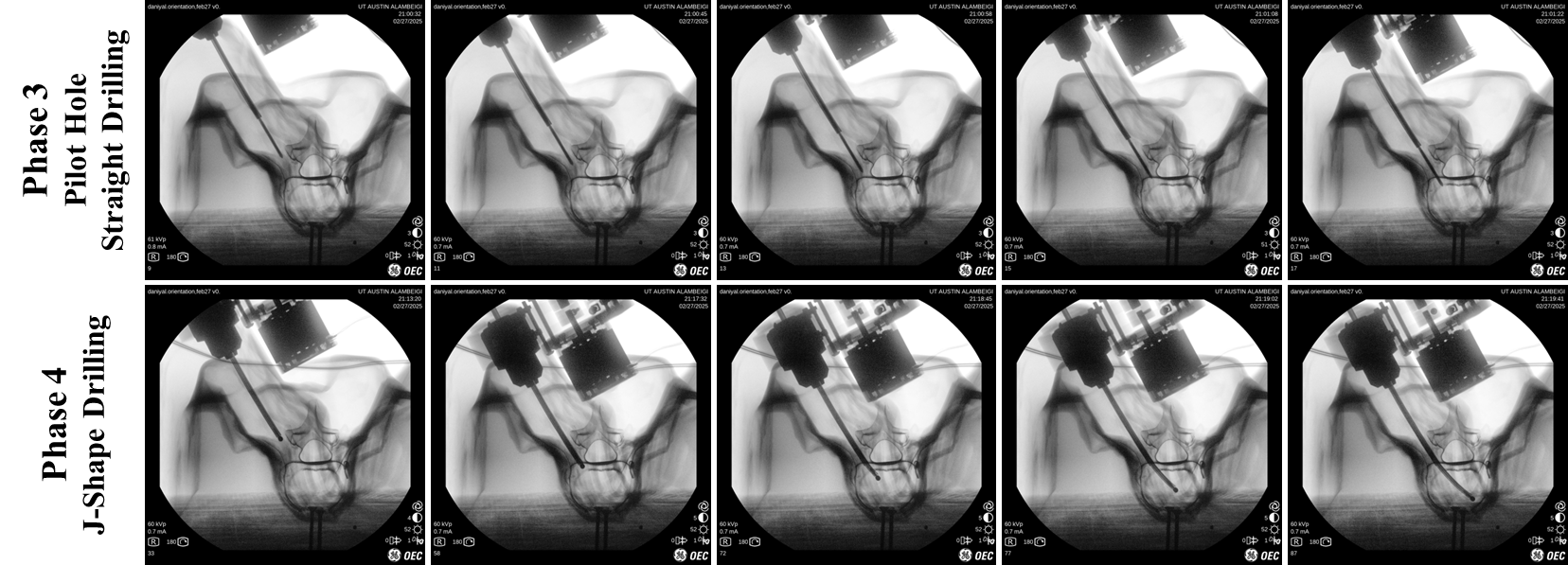}
    \vspace*{-1mm}
	\caption{Sequence of X-Ray scans of Phase 3 and Phase 4 during operation. In Phase 3, the CT-SDR$^{*}$ is used to create a pilot hole using the rigid drilling tool. In Phase 4, the flexible drilling tool is used to create a planar  J-shape trajectory.}
	\label{fig:xray}  
\end{figure*}

\begin{figure}[t] 
	\centering 
    \includegraphics[width=0.95\linewidth]{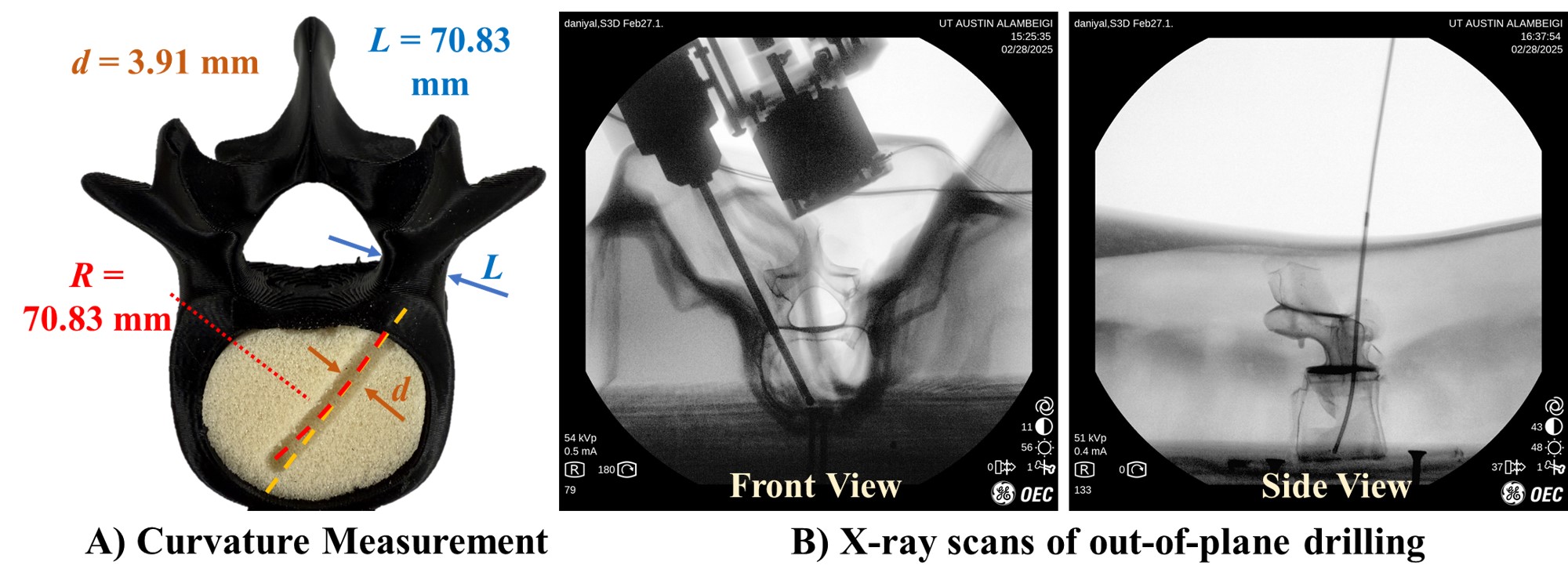}
    \vspace*{-2mm}
	\caption{A) Curvature measurement of the J-shape holes in SOLIDWORKS, B) X-ray scans of the created out-of-plane spatial J-shaped holes and trajectories using the CT-SDR in front and side views.}
	\label{fig:s3d}  
\end{figure}
\subsection{Experimental Setup}

Figure  \ref{fig:setup} shows the experimental setup used to perform the experiments. The setup includes a 7-DoF robotic manipulator (KUKA LBR Med 14, Germany) equipped with the CT-SDR$^{*}$. The setup also included a C-arm X-ray machine (OEC ONE CFD, GE Healthcare) to monitor the CT-SDR$^*$  progress throughout the Phase 3 and Phase 4 drilling procedures. 
Custom-made rigid and flexible tools are used with the CT-SDR$^{*}$ as explained in Section \ref{subsec:CTSDR}.
The flexible steering tube used in this study was a heat-treated, pre-curved NiTi tube with a 69.5 mm radius of curvature. 
An optical tracking tool is attached to the CT-SDR$^*$ to allow tracking by the optical tracking system (NDI Polaris Vega, Northern Digital Inc.). An optical tracking digitizer tool is also utilized to register the desired vertebrae in the framework's coordinate system as explained previously in Section \ref{subsubsec:phase2}. Also, a spine holder back phantom (Spine Holder Posterior T1-Sacrum, Pacific Research Laboratories, USA) is used to simulate the patient's exposed spine. The back phantom secures a custom-designed 3D printed 
L3 vertebral phantom. this phantom is fabricated from ABS material using a Raise3D printer (Raise3D Technologies, Inc.), which includes a space for (i) simulated cancellous bone made from a PCF 5 Sawbone bio-mechanical bone phantom (Pacific Research Laboratories, USA) to mimic a bone with osteoporosis \cite{ccetin2021experimental} and (ii) the cortical bone of vertebral pedicle canal made from a PCF 15 Sawbone.  and house the vertebral phantom.

\subsection{Experimental Procedure} 

\subsubsection{Calibration and Registration Verification}

The hand-eye calibration is verified by moving the robot's end-effector to several randomly selected position and orientation configurations. The actual end effector pose obtained from the robot's forward kinematics is then compared to the calculated end effector pose based on the eye-to-hand calibration results. 
To verify the overall accuracy of the calibration and registration steps- including the eye-to-hand and the digitizer-aided tip calibration- multiple positions and orientations on several vertebrae phantoms were marked using a digitizer tool. The robot was then commanded to move the tool tip to each marked position while maintaining the specified orientation.
Using the transformations obtained from the calibration process, the position and orientation of the tool were calculated and compared with the desired values marked by the digitizer tool. In other words, the pose of the digitizer tip in \{$\text{Polaris}$\} frame when marking the desired pose is compared to the calculated pose of the \{$\text{drill\_tip}$\} frame in  \{$\text{Polaris}$\} frame while the calibration results are used to specify the \{$\text{drill\_tip}$\} frame in \{$\text{Polaris}$\} frame. In this method the transformation between the optical digitizer and its tip is assumed to be accurate. 
The position error is computed as the average Euclidean distance between the two sets of positions. For the orientation error, the difference between the two quaternions is calculated using the conjugate quaternion method, and the result is converted to roll, pitch, and yaw angles to quantify the error. The results are summarized in Table \ref{table:calibrations}. 

\subsubsection{Autonomous Drilling Performance Evaluation}
Performing a complete and realistic drilling towards a spinal fixation surgery scenario is critical in validating the efficacy of both the hardware and software introduced in this framework. Towards this, we used the experimental setup shown in Fig. \ref{fig:setup} to evaluate the proposed S$^{3}$D framework and in particular carry out the four-Phase framework illustrated in Fig. \ref{fig:framework}. 
To make the experiments as realistic as possible, a 3D printed vertebral phantom was secured in placed in a spine phantom, mimicking the anatomy of the exposed spine of the patient in a SF surgery. The CT-SDR$^{*}$ was attached to the end effector of the robotic arm, and a C-arm X-ray machine (OEC One CFD, GE Healthcare) was used to monitor the procedure.
The 3D printed PLA phantom seen in Fig. \ref{fig:setup} had a $7.4\times7.4$ mm hollow hole going through its pedicle with an empty vertebral body. PCF 15 Sawbones (Pacific Research Laboratories, USA) were placed in the pedicle region with PCF 5 Sawbones material being placed in the empty vertebral body.  Then, the phantom is fixed inside the posterior spine phantom, and the entire setup is secured to the desk. Next, Phases 1 (Section \ref{subsubsec:Phase1-calibration}) through 4 (Section \ref{subsubsec:phase4}) of the procedure are performed. after the end of procedure, the Sawbone vertebral phantom is removed for evaluation. The vertebral body is sliced open along the midline of the trajectory, and the cross-section is analyzed using CAD software (SolidWorks, Dassault Systèmes) to measure the curvature of the J-shaped trajectory. We repeated this four-Phase procedure for three different planar  and one out-of-plane drilling trajectories. Figures \ref{fig:xray} and \ref{fig:s3d} show the sequence of  X-ray images for performing planar and out-of-plane spatial drilling trajectories.

\section{Results and Discussion}

\begin{table}[t]
  \centering
 
  \caption{Overall Calibration and Registration Errors}
  \label{table:calibrations}
  \begin{tabular}{|c|c|c|}
    \hline
    \textbf{Error} & \textbf{Rigid Drill Tip} & \textbf{Flexible Drill Tip} \\
    \hline
    \textbf{Position} & \(1.14 \pm 0.28\) mm & \(1.74 \pm 0.97\) mm \\
    \hline
    \textbf{Roll} & \(0.04 \pm 0.04\,\deg\) & \(0.12 \pm 0.20\,\deg\) \\
    \hline
    \textbf{Pitch} & \(0.12 \pm 0.15\,\deg\) & \(0.26 \pm 0.45\,\deg\) \\
    \hline
    \textbf{Yaw} & \(0.20 \pm 0.11\,\deg\) & \(0.80 \pm 1.16\,\deg\)\\
    \hline
   \end{tabular}
\end{table}

As reported in Table \ref{table:calibrations}, an overall positioning error of approximately 1.14 mm and 1.74 mm was achieved for controlling the rigid and flexible drilling tool tip, demonstrating the effective and accurate calibration performance. 
Such position error shows a significant improvement compared to calibration and positioning error in previous studies \cite{sharma2024patient}.
The orientation error was also less than 1 degree, further showing the effectiveness of the calibration methods. Lower calibration error for the rigid drill tip was observed compared to the flexible drilling tool tip. This is due to the sharpness of the rigid drill tip while the flexible drill has a rounded tip which contributes to a slightly higher error during digitizer-aided tip calibration. Given the 3.91 mm diameter of the drilled tunnel and the pedicle canal measuring 12.62 mm (refer to Fig. \ref{fig:s3d}), the obtained calibration error ensures a sufficient safety margin. This margin aligns with the consensus among surgeons that a medical pedicle perforation of less than 4 mm does not pose a risk to adjacent neural structures \cite{Gelalis2012AccuracyOP}.

The X-ray scans of the vertebra inside the phantom during Phase 3 and Phase 4 are shown in Fig. \ref{fig:xray}. During Phase 3, the CT-SDR$^*$ is used to create the pilot hole with the rigid drilling tool, and during Phase 4, the flexible drilling tool is employed to create the J-shaped trajectory. The resulting drilling trajectory and the measured average radius of curvature are depicted in Fig. \ref{fig:s3d}A. With the help of CAD software, we report a radius of curvature of 70.83 $\pm$ 1.72 mm for three tests, representing a 1.9\% error compared to the desired radius of curvature of 69.5 mm. This clearly illustrates the great performance of the proposed $S^3$D framework. While the quantitative analysis is done on J-shape trajectories drilled in the transverse plane of the vertebra, Fig. \ref{fig:s3d}B showcases the system's capability to drill out-of-plane trajectories.

\section{Conclusion and Future Work}

In this paper, we proposed the $S^3$D framework to take a step forward towards performing an effective SF procedure. The introduced  four-Phase calibration, registration, and navigation procedure together with the CT-SDR$^{*}$  enable  drilling in surgeon-specified trajectories to target high BMD regions for pedicle screw placement.
To the best of our knowledge this is the first time to use such realistic experimental setup to perform J-shaped  drilling into the vertebrae. 
We analyzed the precision of the calibration and registration phases by measuring the average position and orientation errors and results showed outstanding performance compared to other state-of-the-art studies. 
Furthermore, realistic drilling scenarios were performed to show the effectiveness of this framework. 
The measured radius of curvature of the resulted holes showed 1.9\% error from the planned trajectory.
In the future we plan to perform animal and cadaver studies to further evaluate the framework in a real SF scenarios. We will also expand the method to control the trajectory plane using an autonomous framework.

\bibliographystyle{./IEEEtran}
\bibliography{./root}

\begin{thebibliography}{10}
\providecommand{\url}[1]{#1}
\csname url@rmstyle\endcsname
\providecommand{\newblock}{\relax}
\providecommand{\bibinfo}[2]{#2}
\providecommand\BIBentrySTDinterwordspacing{\spaceskip=0pt\relax}
\providecommand\BIBentryALTinterwordstretchfactor{4}
\providecommand\BIBentryALTinterwordspacing{\spaceskip=\fontdimen2\font plus
\BIBentryALTinterwordstretchfactor\fontdimen3\font minus \fontdimen4\font\relax}
\providecommand\BIBforeignlanguage[2]{{%
\expandafter\ifx\csname l@#1\endcsname\relax
\typeout{** WARNING: IEEEtran.bst: No hyphenation pattern has been}%
\typeout{** loaded for the language `#1'. Using the pattern for}%
\typeout{** the default language instead.}%
\else
\language=\csname l@#1\endcsname
\fi
#2}}

\bibitem{gaines2000use}
R.~W. Gaines~Jr, ``The use of pedicle-screw internal fixation for the operative treatment of spinal disorders,'' \emph{JBJS}, vol.~82, no.~10, p. 1458, 2000.

\bibitem{rometsch2020screw}
E.~Rometsch, M.~Spruit, J.~E. Zigler, V.~K. Menon, J.~A. Ouellet, C.~Mazel, R.~H{\"a}rtl, K.~Espinoza, and F.~Kandziora, ``Screw-related complications after instrumentation of the osteoporotic spine: a systematic literature review with meta-analysis,'' \emph{Global spine journal}, vol.~10, no.~1, pp. 69--88, 2020.

\bibitem{wittenberg1991importance}
R.~Wittenberg, M.~Shea, D.~Swartz, K.~Lee, A.~White~3rd, and W.~Hayes, ``Importance of bone mineral density in instrumented spine fusions.'' \emph{Spine}, vol.~16, no.~6, pp. 647--652, 1991.

\bibitem{okuyama1993stability}
K.~Okuyama, K.~Sato, E.~Abe, H.~Inaba, Y.~Shimada, and H.~Murai, ``Stability of transpedicle screwing for the osteoporotic spine. an in vitro study of the mechanical stability.'' \emph{Spine}, vol.~18, no.~15, pp. 2240--2245, 1993.

\bibitem{weiser2017insufficient}
L.~Weiser, G.~Huber, K.~Sellenschloh, L.~Viezens, K.~P{\"u}schel, M.~M. Morlock, and W.~Lehmann, ``Insufficient stability of pedicle screws in osteoporotic vertebrae: biomechanical correlation of bone mineral density and pedicle screw fixation strength,'' \emph{European Spine Journal}, vol.~26, no.~11, pp. 2891--2897, 2017.

\bibitem{Li2023RoboticSA}
T.~Li, A.~Badre, F.~Alambeigi, and M.~Tavakoli, ``Robotic systems and navigation techniques in orthopedics: A historical review,'' \emph{Applied Sciences}, 2023.

\bibitem{CT_MRI_registration}
B.~Jian, M.~F. Azampour, F.~De~Benetti, J.~Oberreuter, C.~Bukas, A.~S. Gersing, S.~C. Foreman, A.-S. Dietrich, J.~Rischewski, J.~S. Kirschke, N.~Navab, and T.~Wendler, ``Weakly-supervised biomechanically-constrained ct/mri registration of the spine,'' in \emph{Medical Image Computing and Computer Assisted Intervention -- MICCAI 2022}, L.~Wang, Q.~Dou, P.~T. Fletcher, S.~Speidel, and S.~Li, Eds.\hskip 1em plus 0.5em minus 0.4em\relax Cham: Springer Nature Switzerland, 2022, pp. 227--236.

\bibitem{stent_recovery}
S.~Demirci, A.~Bigdelou, L.~Wang, C.~Wachinger, M.~Baust, R.~Tibrewal, R.~Ghotbi, H.-H. Eckstein, and N.~Navab, ``3d stent recovery from one x-ray projection,'' in \emph{Medical Image Computing and Computer-Assisted Intervention -- MICCAI 2011}, G.~Fichtinger, A.~Martel, and T.~Peters, Eds.\hskip 1em plus 0.5em minus 0.4em\relax Berlin, Heidelberg: Springer Berlin Heidelberg, 2011, pp. 178--185.

\bibitem{ElmiTerander2018PedicleSP}
A.~Elmi-Terander, G.~Burstr{\"o}m, R.~Nachabe, H.~Sk{\'u}lason, K.~Pedersen, M.~Fagerlund, F.~St{\aa}hl, A.~Charalampidis, M.~S{\"o}derman, S.~Holmin, D.~Babic, I.~Jenniskens, E.~Edstr{\"o}m, and P.~Gerdhem, ``Pedicle screw placement using augmented reality surgical navigation with intraoperative 3d imaging,'' \emph{Spine}, vol.~44, pp. 517 -- 525, 2018.

\bibitem{beyond2025Bhimreddy}
\BIBentryALTinterwordspacing
M.~Bhimreddy, A.~K. Menta, A.~A. Fuleihan, A.~D. Davidar, P.~Kramer, R.~Jillala, M.~Najeed, X.~Wang, and N.~Theodore, ``Beyond pedicle screw placement: Future minimally invasive applications of robotics in spine surgery,'' \emph{Neurosurgery}, vol.~96, no.~3S, 2025. [Online]. Available: \url{https://doi.org/10.1227/neu.0000000000003335}
\BIBentrySTDinterwordspacing

\bibitem{wang2021design}
Y.~Wang, H.-W. Yip, H.~Zheng, H.~Lin, R.~Taylor, and K.~W.~S. Au, ``Design and experimental validation of a miniaturized robotic tendon-driven articulated surgical drill for enhancing distal dexterity in minimally invasive spine fusion,'' \emph{IEEE/ASME Transactions on Mechatronics}, 2021.

\bibitem{9732206}
Y.~Wang, H.~Zheng, R.~H. Taylor, and K.~W.~S. Au, ``A handheld steerable surgical drill with a novel miniaturized articulated joint module for dexterous confined-space bone work,'' \emph{IEEE Transactions on Biomedical Engineering}, pp. 1--1, 2022.

\bibitem{alambeigi2019use}
F.~Alambeigi, M.~Bakhtiarinejad, S.~Sefati, R.~Hegeman, I.~Iordachita, H.~Khanuja, and M.~Armand, ``On the use of a continuum manipulator and a bendable medical screw for minimally invasive interventions in orthopedic surgery,'' \emph{IEEE transactions on medical robotics and bionics}, vol.~1, no.~1, pp. 14--21, 2019.

\bibitem{alambeigi2017curved}
F.~Alambeigi, Y.~Wang, S.~Sefati, C.~Gao, R.~J. Murphy, I.~Iordachita, R.~H. Taylor, H.~Khanuja, and M.~Armand, ``A curved-drilling approach in core decompression of the femoral head osteonecrosis using a continuum manipulator,'' \emph{IEEE Robotics and Automation Letters}, vol.~2, no.~3, pp. 1480--1487, 2017.

\bibitem{sharma2024patient}
S.~Sharma, S.~Go, Z.~Yakay, Y.~Kulkarni, S.~Kapuria, J.~P. Amadio, R.~Rajebi, M.~Khadem, N.~Navab, and F.~Alambeigi, ``A patient-specific framework for autonomous spinal fixation via a steerable drilling robot,'' in \emph{International Conference on Medical Image Computing and Computer-Assisted Intervention}.\hskip 1em plus 0.5em minus 0.4em\relax Springer, 2024, pp. 35--45.

\bibitem{Sharma_tbme_2022}
S.~Sharma, T.~G. Mohanraj, J.~P. Amadio, M.~Khadem, and F.~Alambeigi, ``A concentric tube steerable drilling robot for minimally invasive spinal fixation of osteoporotic vertebrae,'' \emph{IEEE Transactions on Biomedical Engineering}, vol.~70, no.~11, pp. 3017--3027, 2023.

\bibitem{Sharma_ismr}
S.~Sharma, Y.~Sun, S.~Go, J.~P. Amadio, M.~Khadem, A.~H. Eskandari, and F.~Alambeigi, ``Towards biomechanics-aware design of a steerable drilling robot for spinal fixation procedures with flexible pedicle screws,'' in \emph{2023 International Symposium on Medical Robotics (ISMR)}, 2023, pp. 1--6.

\bibitem{Kulkarni2025TowardsDD}
Y.~Kulkarni, S.~Sharma, S.~Go, J.~P. Amadio, M.~Khadem, and F.~Alambeigi, ``Towards design and development of a concentric tube steerable drilling robot for creating s-shape tunnels for pelvic fixation procedures,'' \emph{ArXiv}, 2025.

\bibitem{Kulkarni2025SynergisticPSA}
Y.~Kulkarni, S.~Sharma, Z.~Yakay, S.~Go, J.~P. Amadio, M.~Tilton, and F.~Alambeigi, ``A synergistic patient-specific approach for enhanced spinal fixation using a novel flexible pedicle screw and a complementary steerable drilling robotic system,'' \emph{IEEE Transactions on Biomedical Engineering}, pp. 1--11, 2025.

\bibitem{Kulkarni2024SFF}
Y.~Kulkarni, S.~Sharma, J.~Allison, J.~P. Amadio, M.~Tilton, and F.~Alambeigi, ``Towards the feasibility analysis and additive manufacturing of a novel flexible pedicle screw for spinal fixation procedures,'' \emph{The 35th Annual International Solid Freeform Fabrication Symposium}, pp. 1--11, 2024.

\bibitem{Kulkarni2025AugmentedBSF}
Y.~Kulkarni, S.~Sharma, O.~Rezayof, S.~Kapuria, J.~P. Amadio, M.~Khadem, M.~Tilton, and F.~Alambeigi, ``Augmented bridge spinal fixation: A new concept for addressing pedicle screw pullout via a steerable drilling robot and flexible pedicle screws,'' \emph{ArXiv}, 2025.

\bibitem{sharma2023novel}
S.~Sharma, J.~H. Park, J.~P. Amadio, M.~Khadem, and F.~Alambeigi, ``A novel concentric tube steerable drilling robot for minimally invasive treatment of spinal tumors using cavity and u-shape drilling techniques,'' in \emph{2023 IEEE International Conference on Robotics and Automation (ICRA)}.\hskip 1em plus 0.5em minus 0.4em\relax IEEE, 2023, pp. 4710--4716.

\bibitem{sharma2024biomechanics}
S.~Sharma, Y.~Sun, J.~Bonyun, M.~Khadem, J.~Amadio, A.~H. Eskandari, and F.~Alambeigi, ``A biomechanics-aware robot-assisted steerable drilling framework for minimally invasive spinal fixation procedures,'' \emph{IEEE Transactions on Biomedical Engineering}, vol.~71, no.~6, pp. 1810--1819, 2024.

\bibitem{shah2013solving}
M.~Shah, ``Solving the robot-world/hand-eye calibration problem using the kronecker product,'' \emph{Journal of Mechanisms and Robotics}, vol.~5, no.~3, p. 031007, 2013.

\bibitem{opencv_library}
G.~Bradski, ``{The OpenCV Library},'' \emph{Dr. Dobb's Journal of Software Tools}, 2000.

\bibitem{yaniv2015pivot}
Z.~Yaniv, ``Which pivot calibration?'' in \emph{Medical imaging 2015: Image-guided procedures, robotic interventions, and modeling}, vol. 9415.\hskip 1em plus 0.5em minus 0.4em\relax SPIE, 2015, pp. 542--550.

\bibitem{ccetin2021experimental}
A.~{\c{C}}etin and D.~A. Bircan, ``Experimental investigation of pull-out performance of pedicle screws at different polyurethane (pu) foam densities,'' \emph{Proceedings of the Institution of Mechanical Engineers, Part H: Journal of Engineering in Medicine}, vol. 235, no.~6, pp. 709--716, 2021.

\bibitem{Gelalis2012AccuracyOP}
I.~Gelalis, N.~K. Paschos, E.~E. Pakos, A.~N. Politis, C.~M. Arnaoutoglou, A.~C. Karageorgos, A.~Ploumis, and T.~A. Xenakis, ``Accuracy of pedicle screw placement: a systematic review of prospective in vivo studies comparing free hand, fluoroscopy guidance and navigation techniques,'' \emph{European Spine Journal}, vol.~21, pp. 247--255, 2012.

\end{thebibliography}

\end{document}